\documentclass[twoside]{article}
\usepackage{PRIMEarxiv}

\usepackage{times}
\usepackage{soul}
\usepackage{url}
\usepackage[hidelinks]{hyperref}
\usepackage[utf8]{inputenc}
\usepackage[small]{caption}
\usepackage{graphicx}
\usepackage{amsmath}
\usepackage{amsthm}
\usepackage{booktabs}
\usepackage{algorithm}
\usepackage{algorithmic}
\usepackage[switch]{lineno}

\usepackage{xspace}
\usepackage{amsmath}
\usepackage{sty/myeditingtools}
\usepackage{sty/mynotations}
\usepackage{svg}
\usepackage{tablefootnote}

\title{Constraints First: \\ A New MDD-based Model to Generate Sentences Under Constraints}

\author{
    Alexandre Bonlarron \\
    Université Côte d’Azur, Inria, France \\
    Université Côte d’Azur, CNRS, I3S, France \\
    alexandre.bonlarron@inria.fr \\
    \and
    Aurélie Calabrèse \\
    Université Aix-Marseille, CNRS, LPC, France \\
    aurelie.calabrese@univ-amu.fr
    \and
    Pierre Kornprobst \\
    Université Côte d’Azur, Inria, France \\
    pierre.kornprobst@inria.fr
    \and
    Jean-Charles Régin \\
    Université Côte d’Azur, CNRS, I3S, France\\
    jean-charles.regin@univ-cotedazur.fr
}
\date{} % Leave this blank or specify a date

\begin{document}

\maketitle

\begin{abstract}

This paper introduces a new approach to generating strongly constrained texts. We consider standardized sentence generation for the typical application of vision screening. To solve this problem, we formalize it as a discrete combinatorial optimization problem and utilize multivalued decision diagrams (MDD), a well-known data structure to deal with constraints. In our context, one key strength of MDD is to compute an exhaustive set of solutions without performing any search. Once the sentences are obtained, we apply a language model (GPT-2) to keep the best ones. We detail this for English and also for French where the agreement and conjugation rules are known to be more complex. Finally, with the help of GPT-2, we get hundreds of bona-fide candidate sentences. When compared with the few dozen sentences usually available in the well-known vision screening test (MNREAD), this brings a major breakthrough in the field of standardized sentence generation. Also, as it can be easily adapted for other languages, it has the potential to make the MNREAD test even more valuable and usable. More generally, this paper highlights MDD as a convincing alternative for constrained text generation, especially when the constraints are hard to satisfy, but also for many other prospects.

\end{abstract}

\section{Introduction}
\label{sec:introduction}

We consider the problem of generating text under constraints.
Currently, the methodology for solving this problem is based on two main components:
(a) generating words using Language Models (LM) that have been proven on several tasks in natural language processing~\cite{glue:2018}\footnote{https://gluebenchmark.com/},
(b) processing text sequentially to satisfy constraints which can be done by tokens (words) filtering~\cite{roush-etal-2022-language} or search mechanisms such as Beam Search (BS)~\cite{beamsearch1:2021,beamsearch2:2018,hokamp-liu:2017} or more recently A*~\cite{lu-etal-2022-neurologic}.
These search-based heuristics involve generating the text word-by-word while maximizing the likelihood of the sequence computed by (a), exploring solutions space, and checking constraints with (b). 
This methodology is appropriate as long as the solution space is \emph{weakly constrained}, i.e. when it is easy to find a sequence satisfying the constraints. This is usually the case for lexical or semantic constraints since they are local requirements in the output and are almost formalized as output preferences.

%More precisely, it relies on large-scale LMs like GPT \cite{radford:2019} link to 
%These methods aim to deal with lexical or semantic constraints, ensuring that a particular list of words or concepts are in the output. More recently, a new method was proposed in the same spirit to use A* instead of a BS \cite{lu-etal-2022-neurologic}. Sometimes the lexical constraint can be handled without a search and uses tokens (words) filtering \cite{roush-etal-2022-language}. 

In this paper, we are interested in generating constrained text when the solutions space is \emph{strongly constrained} (i.e., when it is hard to find a solution that satisfies the constraints). 
This is the case when constraints are not local anymore (e.g., length constraint, display constraint). In other words, they are defined on the whole text.
Because the sequences we are looking for are rare, we propose a completely different methodology based on three ideas:
(i) We propose to check the constraints first and then select the best solutions with an LM because there is no reason that the sequences that an LM considers likely would satisfy the constraint first.
(ii) We see a constrained text generation problem as a discrete combinatorial optimization problem, where variables are words, the domain of variables is the vocabulary of text, and constraints are the set of rules the text has to satisfy. 
(iii) %We use n-grams to take into account language structure.
%This formulation generalizes a similar idea developed for generating palindromes (e.g., race car)~\cite{papadopoulos:2015}.
Even though we want to satisfy the constraints first, applying a pure {\itshape{generate and test}} over random words would find sequences satisfying the constraints, but those sequences would not have a meaning. In other words, these sequences are not "well-formed" sentences. Since we cannot totally ignore the language structure, we rely on combining existing sequences of words (n-grams). For example, from the sentence: "He lives in a nice little red house", we can ﬁnd
several 3-grams (e.g., "a nice little", "nice little red", "little red house"). On top of that, we can link two n-grams if the $n-1$ last words of the first \ngram are the $n-1$ first words of the second \ngram. Then sentences are obtained by following links between n-grams. Note that this idea of combining n-grams was used to generate palindromes (e.g., race car)~\cite{papadopoulos:2015} with an ad-hoc algorithm.

%Papadopolous et al. start from ngram to build the greatest palindromes \cite{papadopoulos-roy-etal:14}
%This paper proposes a Constraint Programming (CP) model for solving a constrained text generation problem and shows its effectiveness through the MNREAD sentences generation problem. The MNREAD sentences generation problem produces standardized sentences satisfying a sharp protocol (i.e., a set of strict rules). These sentences are used to build the MNREAD reading test \cite{mansfield-ahn-etal:93}, which is a well-known test in the field of vision research for measuring reading speed. Increasing the number of MNREAD sentences is a complex problem because of the several hard constraints the sentences must follow. Writing these sentences or ﬁnding them in corpora is not the solution because they are particular; so particular that almost no existing instances satisfy all the rules. Out of a sample of more than three million sentences from children's literature (600 books), only four satisfy the criteria of the MNREAD reading test.

Our contribution is twofold: we propose a Constraint Programming (CP) model for sentence generation and apply our model to generate standardized sentences for the MNREAD test~\cite{mansfield-ahn-etal:93}, a well-known test in vision research for measuring reading performance. MNREAD sentences must satisfy a sharp protocol (i.e., a set of strict rules), and they are hard to find. To be convinced, out of a sample of more than three million sentences from children's literature (600 books), we only found four obeying all rules.

%sentences are short sentences that should be very simple to read and all equivalent (in the sense that they should all be read at the same speed). In practice, this notion of equivalence which is fundamental to avoid any bias, is inferred by a set of rules that will make these sentences special although at first glance they might seem trivial. 

Our approach is based on multi-valued decision diagrams (MDDs) to produce sentences. An MDD is a well-suited data structure to store and retrieve successions of \ngrams. Also, using MDD provides a way to compute an exhaustive set of solutions without performing a search. Therefore, our approach is an exact method.
% XXX from which the best ones are selected with Transformers.
In addition, we use a Language Model (LM) to select the best sentences.
An LM is a probabilistic model used to predict the sequence of words in a text or speech using natural language training data. Here we use transformers-based LM (i.e., LM implementing Transformers architecture based on attention block \cite{vasmani-et-al:2017}). More specifically, we use GPT-2 \cite{radford:2019}, which has been trained over a tremendous amount of text and is designed to generate sequences. GPT-2 is also suited to assess sentence quality to a certain degree by assigning a score to sequences. We use this score to sort sentences and select the best ones. It drastically simplifies human verification and, in the best case, avoids it.

The paper is organized as follows: we give some preliminaries in Sec. \ref{sec:prelim}, with a presentation of the MNREAD rules and MDDs.
Then in Sec. \ref{sec:methode}, we show how to model the constrained text generation problem with MDDs.
We also explain how to use an LM to select the best sentences.
The experimental results are presented in Sec. \ref{sec:resultats}, in which we show the potential of our method.
At last, in Sec. \ref{sec:discussion}, we share additional thoughts on this work and give some perspectives on future investigations.
Finally, we conclude.

\section{Preliminaries}
\label{sec:prelim}
\subsection{Case Study: MNREAD Test}

\subsubsection{Motivation}
Reading performance has become one of the most widely used clinical measures to judge the effectiveness of treatments, surgical procedures, or rehabilitation techniques. This reading performance is measured by the time taken to read standardized texts, i.e., texts designed to be equivalent in terms of length, display, and linguistics. This need for equivalent texts is the key to avoid bias in the texts themselves. In addition, each text should only be presented once to the subjects or patients to avoid recall bias and to ensure an accurate measurement.

Among various existing tests, the MNREAD test~\cite{mansfield-ahn-etal:93} is probably one of the most widely used standardized reading tests in the world for measuring reading performance in clinical and research settings in people with normal or low vision.
For instance, it is used to evaluate how reading performance depends on font size~\cite{calabrese-owsley-etal:16}. One MNREAD test comprises 19 standardized sentences printed in decreasing sizes. 
It is available in 19 languages, with different numbers of tests depending on the language (e.g., five in English, two in French). Since repeated measurements are needed, a large number of test sets are required. Unfortunately, currently, the number of available MNREAD sentences is largely insufficient. 

\subsubsection{Related Work}
Some researchers designed ad-hoc approaches dedicated to reading tests~\cite{crossland-legge-etal:07,perrin-paille-etal:15,mansfield-atilgan-etal:19}. In particular, Mansfield et al~\cite{mansfield-atilgan-etal:19}, who are also the creators of the MNREAD test, proposed an approach based on sentence templates capable of generating millions of sentences using a random walk. 
However, this semiautomatic method has four major drawbacks: (i) It relies on sentence templates that have
to be created manually (i.e., sequences of placeholders, each containing
a list of possible words that ﬁt into the sentence at a deﬁned
position). (ii) It requires a manual verification of the sentences generated by the system. (iii) It cannot be easily extended 
to other languages, such as French, where one has to take into
account agreements and conjugations or other languages that use
declension (e.g., German, Slavic, ...). (iv) Since the
sentences generated are based on templates, the number of different
words is limited. It may lead to a memory bias for readers.

\subsubsection{Characterisation of MNREAD Sentences: Definition of Rules}
\label{sec:mnread-regles}
\begin{figure}[t]
    \centerline{\includegraphics[width=\textwidth]{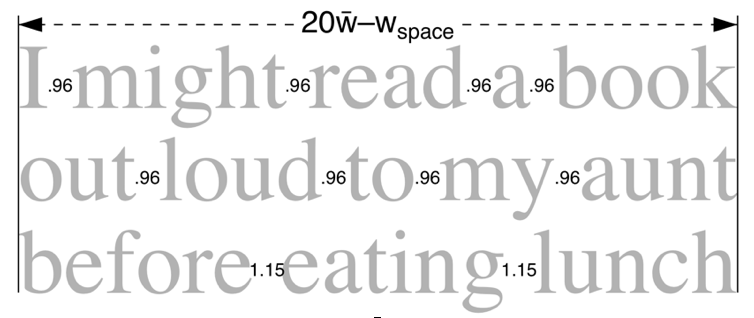}}
   \caption{\label{fig:mnread-sentence}
   An MNREAD sentence that satisfies the constraints $\regleGrammaticale\ldots\regleAffichage$. The sentence is displayed with left and right justification on three lines of text and must fit in a box whose width is 17.3 times the size of a standard Times-Roman letter. The width of the spaces between the words (shown as a multiple of the normal width of the spaces) should be between 0.80 and 1.25. %Courtesy~ \cite{mansfield-atilgan-etal:19} }
   }
\end{figure}

In order to be standardized, all MNREAD sentences must satisfy five rules $(\Rule{i})_{i=0..4}$ made explicit by the creators of the test (see~\cite{mansfield-atilgan-etal:19} for more details): 
\begin{compactitemize}
\item ($\regleGrammaticale$) grammatical rules: sentences in declarative form, no punctuation, no proper nouns 
\item $(\regleLexique)$ lexical rules: sentences constructed only from the 3000 lemmas
(canonical form) of the most frequent words in the 3rd-grade level
textbooks, as deﬁned in the Manulex~\cite{lete-sprenger-charolles-etal:04} lexicon
\item $(\regleNombreMots, \regleNombreCaracteres)$ length rules: $(\regleNombreMots)$ sentences having from 9 to 15 words and $(\regleNombreCaracteres)$ 59 characters with spaces and without counting the period  
\item ($\regleAffichage$) display rules: sentences that can be displayed on three lines, with left-right justification and inter-word spaces between strict threshold values, proportional to the size of the common space for the given font
\end{compactitemize}
An example of an MNREAD sentence respecting these rules is given in Fig.~\ref{fig:mnread-sentence}  (adapted from~\cite{mansfield-atilgan-etal:19}).

The rules $(\Rule{i})_{i=0..4}$ are necessary but not sufficient to belong to the test.
Indeed, there are also implicit rules, more qualitative and more challenging to formalize but
which seem ``obvious'', such as semantic rules (the sentence must have a meaning) or syntactic rules (the sentence
must have a ``simple'' structure). 
Thus, a sentence will be considered of type MNREAD if
it complies with all the rules previously stated,
%A sentence considered of type MNREAD will thus be a sentence that respects all the rules previously stated,
makes sense and ﬁnally, resembles the ofﬁcial sentences.

\subsection{Multi-Valued Decision Diagram}

MDDs are a generalization of binary decision diagrams (BDDs)~\cite{akers:78}. 
MDDs have already been successfully used in many other problems, such as the generation of music~\cite{roy-perez-etal:16} or poems~\cite{perez-regin:17b}. MDDs are usually a background data structure in CP solver initially introduced for table constraints~\cite{cheng-yap:10,lecoutre:2011} and are sufficiently generic to compute and store almost any constraint  \cite{Wang-Yap:2022,gentzel-etal:2022,verhaeghe2019extending,verhaeghe2018compact}. Other uses of MDDs for optimization problems can be found in ~\cite{bergman-cire-etal:16,perez:17,gillard2022large,rudich:2022peelandbound,Hoeve:2022}.

They are data structures for computing and storing solutions (tuples) of a problem using a directed acyclic graph (DAG).
MDDs allow representing a large number of solutions in a compressed structure. 

Using MDDs in our context offers three main advantages: 
(i) MDD is a straightforward extension and an efﬁcient way of compactly representing a corpus (i.e., the succession of words/n-grams of a text).
(ii) MDD is a very powerful modeling tool.
(iii) It allows the computation of an exhaustive set of solutions without performing a search.

In this paper, we consider deterministic reduced ordered MDDs \cite{Amilhastre:2014}. MDDs are structured in layers where each layer represents an ordered variable $X_i$.
There are two particular nodes in the MDD: the root (\nroot) and the true terminal node (\ntt). Each path between the node \nroot and \ntt forms a label tuple corresponding to an assignment of the variables associated with each layer. 
%The label $l_{a_{k}}$ of the arc $a_k$ belonging to any path between the layer $i-1$ and $i$ is the assignment of the variable $X_i$ such that $X_i$ = $l_{{a}_k}$. 
More precisely, the label $w_{a_{k}}$ of the arc $a_k$ for the variable $X_i$ means that the variable $X_i$ is assigned to $w_{{a}_k}$, and  a path $p$ = ($a_1$, $a_2$, $a_3$, $\ldots$, $a_r$) corresponds to the global assignment ($w_{a_1}$, $w_{a_2}$, $w_{a_3}$, $\ldots$, $w_{a_r}$) of the variables (i.e. an $r$-uplet, or a in our context a sentence of $r$ words).

%A path $p$ = ($a_1$, $a_2$, $a_3$, $\ldots$, $a_r$) exists in the MDD if and only if the variable assignment $X_i$=$l_{a_{i}}$ is valid for all $i$. 
%Thus computing an MDD containing $r$ layers is equivalent to calculating the valid $r$-uplets of the function $f : \{0...d\}^r \mapsto \{true,false\}$.

%\paragraph{Language Model (LM): }
%A language model is a probabilistic model used to predict the sequence of words in a text or speech using natural language training data. Language models are often used to improve text comprehension, spelling correction, text generation, machine translation, and other tasks related to NLP. They work by learning the likelihoods of different combinations of words in a text to predict the most likely sequence of words in a new text. The best current LMs in NLP are transformers-based models like GPT-2.

%\subsection{Perplexity Measure (PPL)} 
%It is a cross-entropy measure derived from Shannon's information theory. \ab{ cite Brown et al (entropy) 1992}
%The perplexity can be expressed as the geometric mean of the inverse conditional likelihood of the sequence; let the sequence $S_n$ a succession of words $S_n = w_1w_2..w_n$ then PPL of $S_n$ is computed as follow :
%\[ PPL(S_n) = \sqrt[n]{\frac{1}{P(w_1w_2w_3...w_n)}} \]
%The probability $P(...)$ is given by the LM. 
%This measure can be viewed as : the "uncertainty" of the model with respect to a sample.
%Usually, it is used to evaluate the LM itself by choosing good samples and validating that the good samples are recognised as such (i.e., low PPL values)

\section{Our Approach}\label{sec:approach}
\label{sec:methode}

First, we define our input data (i.e., an \ngrams set). Then, we explain how to efficiently manipulate these n-grams using MDDtrie, a specific MDD.
Next, we describe how to compile the final MDDMNREAD thanks to the MDDTrie to obtain the solution sets (i.e., the generated sentences).
Finally, we explain how we use an LM to sort sentences.

\subsection{Definition of the Input}
%In practice, as is classically done, we will create a list of \ngrams from a corpus consisting of a set of books. 
Let us consider a set of \ngrams, extracted from sentences gathered in a corpus (a set of books).
The specificity of our application is that we will ignore a certain number of sentences (and therefore of \ngrams) in order to integrate the $\Rule{0}$ grammatical rules from the beginning (e.g., interrogative sentences will be ignored, in order not to add \ngrams corresponding to interrogative turns of phrases which are not accepted as MNREAD sentences). %Then, a series of transformations using regular expressions are applied to the sentences of the corpus to obtain more usable \ngrams (e.g., ellipsis, colon, and semicolon are considered full stops; inverted commas are ignored). 
Finally, thanks to the part-of-speech tagging feature of treetagger library \cite{schmid:13}, we filter out any \ngram that does not fit the lexical rule ($\regleLexique$) and also punctuations (e.g., ellipsis, colon, and semicolon and inverted commas). 
In addition, we differentiate between three types of \ngrams: initial \ngrams (starting a sentence), middle \ngrams, and final \ngrams (ending a sentence). In this way, depending on the position of the word in the sentence we are trying to assign, some \ngrams are allowed and others are not.
We explain in the following sections how to recombine them while satisfying constraints.
%The specificity of our problem is that we will ignore a certain number of sentences (and therefore of \ngrams) in order to integrate the $\Rule{0}$ grammatical rules from the beginning (e.g., interrogative sentences will be ignored, in order not to add \ngrams corresponding to interrogative turns of phrase which are not desired in our application). Then, a series of transformations using regular expressions are applied to the sentences of the corpus to obtain more usable \ngrams (e.g., ellipsis, colon, and semicolon are considered full stops; inverted commas are ignored). Next, some rewrites are applied to standard status abbreviations to recover their unabbreviated forms (e.g., Mr. is rewritten Mister). Finally, we filter out any \ngram that does not fit the vocabulary rule $\regleLexique$. 

\subsection{Defining MDDs Associated With Constraints}

\subsubsection{MDDtrie: the MDD of the Succession Constraint}

The goal of MDDtrie is to define a succession rule between words.
To store and reTRIEve our \ngrams and their successors efficiently,
MDDtrie contains all \ngrams of a fixed length. In order to build it, we insert all \ngrams as paths (i.e., solutions) and apply the reduction operator (i.e., pReduce, \cite{perez-regin:15}). 
As a result of MDD properties, each \ngram that shares the same prefix of words shares the same prefix path from \nroot to \ntt.

Given that we can link two \ngrams if the $n-1$ last words of the first \ngram are the $n-1$ first words of the second \ngram,
 finding all words successors of an \ngram $w_1w_2w_3...w_n$ is equivalent to searching in MDDtrie the list of labels of outgoing arcs of the node that can be reached from the \nroot with the subpath  $(w_2,w_3,...,w_n)$ (i.e., the suffix of size $n-1$). 

MDDtrie is a different way of defining a gliding constraint with an MDD (see for instance \cite{jung-regin:22} in the case of the sequence constraint ).

\begin{figure}[htbp]
    \centering
    \includesvg[width=0.65\textwidth]{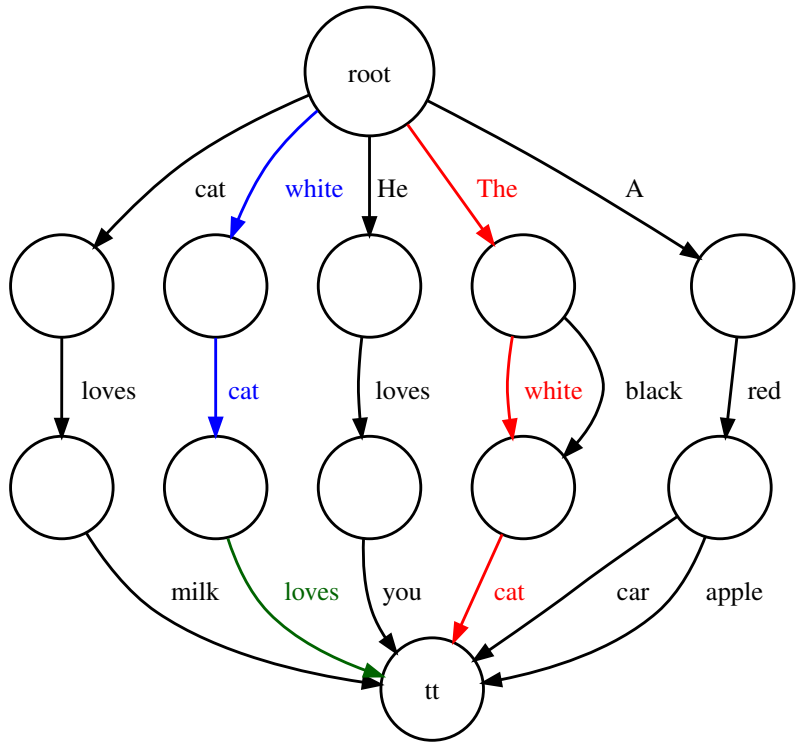}
    \caption{Example of MDDtrie storing 3-grams (successions of 3 words): "The black cat"; "A red apple"...
    Any path from the \nroot to \ntt is a valid \ngram. 
%    In red, the \ngram "{\color{red}The white cat"} is highlighted. To find the successors of this \ngram, more precisely the following potential words,
    To find the successors of the \ngram "{\color{red}The white cat}", more precisely the following potential words,
 we start from \nroot a walk along the arcs that contains the labels of the two last arc, i.e., "{\color{blue}white}" and "{\color{blue}cat}".    
%    start from \nroot with the labels of the two last arc, i.e., "{\color{blue}white}" and "{\color{blue}cat}". We start from \nroot a walk along the arcs that contains these labels (in blue). 
In that case, one outgoing arc from the node can be reached with "white cat". Thus, the successor of "The white cat" is "{\color{darkgreen}loves}" (in green).}
%    MDDtrie allows the storage and retrieval of \ngrams and their successors efficiently.}
    \label{fig:MDDtrie}
\end{figure}

%There is a strong relationship between MDD and a Trie \cite{briandais:1959}. 
An example of the use of MDDtrie is illustrated in a simplified case in Fig.~\ref{fig:MDDtrie}.

%This MDD, called MDDTrie, can be used as a Trie \cite{briandais:1959}.

\subsubsection{MDDMNREAD: The MDD of the Rules $\regleNombreMots \cap \regleNombreCaracteres \cap \regleAffichage$
}

\begin{figure}[htbp]
 \centering
  \includesvg[width=0.8\textwidth]{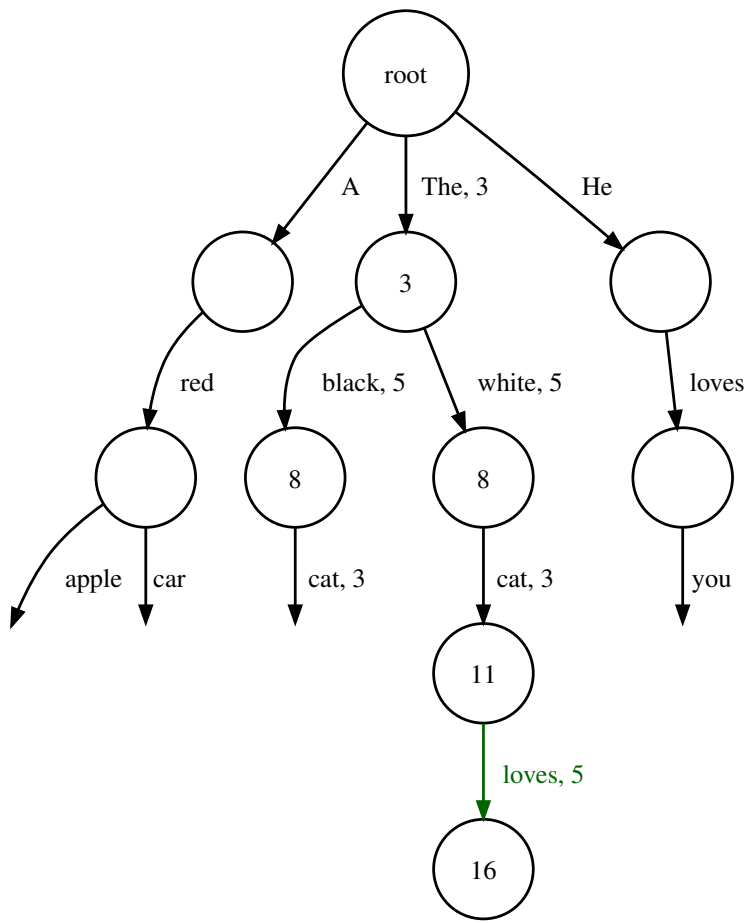}
  \caption{\label{fig:illustrations-unfolding}
  Unfolding of MDDMNREAD with four layers of nodes using the MDDtrie of Fig. \ref{fig:MDDtrie}. We focus our explanation on the path "the white cat loves", the successor "loves" (in green) was found using MDDtrie. While expanding MDDtrie, we compute on the fly the cost of each label (here, the number of characters) and as long as the cost satisfies the constraints ($\regleNombreCaracteres$) we keep expanding using the succession rule. 
Other constraints, like the display one ($\regleAffichage$), can be computed in the same fashion at the same time.}
%  Illustrations of  unfolding of MDDMNREAD in a simplified framework with four layers of nodes using MDDtrie in figure \ref{fig:MDDtrie}: For readability reasons we focus our explanation on only the path "the white cat loves", the successors "loves" (in green) was found using MDDtrie. While expanding MDDtrie, we compute on the fly the cost of each labels (here the length i.e., the number of characters) and while the cost satisfying the constraints ($\regleNombreCaracteres$) 
%  we keep expanding using the Succession Rule. 
  %The information is spread along layers using nodes since MDD are also suited to Dynamic Programming computing
  %\cite{trick:03}. }
 % Obviously, the other constraint like the display one ($\regleAffichage$) is computed in the same fashion at the same time,
 %and if we were computing the cost for the display constraint, we reset the sum associated to ($\regleAffichage$) after the label loves. (since it make the first line valid and full).}
\end{figure}

This new MDD is an unfolded MDD of the MDDTrie of size 15 ($\regleNombreMots$) . While unfolding it (i.e, applying the succession rule), we solve the constraint problem induced by MNREAD sentence rules (see Fig. \ref{fig:illustrations-unfolding}). 

It will contain all possible combinations of words and indirectly of n-grams leading to the production of sentences. 
The n-grams of the ﬁrst (resp. last) layer must correspond to the n-grams of the start (resp. end) of the sentence. This eliminates sentences that do not end correctly by avoiding ending on a word that is supposed to introduce other words. As a result, some words are defined as non-terminal (e.g., and, or, to...). The empty word is only possible if the node \ntt or an empty word follows it. This trick makes it possible to build an MDD containing sentences of different sizes. 

In the following paragraphs, while expanding MDDMNREAD, we describe how to compute the remaining constraints.
Keep in mind that the label of arcs of the MDDMNREAD is always a {\itshape{word}}. We also associate costs with each arc. A cost is computed on the fly in relation to the considered constraint (e.g., length, size of the word).
So MDDMNREAD is also virtually a cost-MDD.

For constraint ($\regleNombreCaracteres$) : we ensure that each sentence has 59 characters, including spaces by
imposing a sum constraint \cite{trick:03} on the length of sentences.
In this case, the cost of an arc is the number of characters in a word.
Then, MDDMNREAD could be seen as a particular sum MDD whose sum of the labels of the arcs that constitute any solution is equal to 59. (see in Fig. \ref{fig:illustrations-unfolding} where cost is associated with the label, and the partial sum is propagated through nodes).

For constraint ($\regleAffichage$) : We ensure the display rule (see example in Fig. \ref{fig:mnread-sentence}), which states that a sentence must be displayed on three lines, by imposing a sum constraint on the width of the characters of the sentence. The width of a character is defined according to the font used.
%In addition, all the spaces of a line must have the same width $e$, and this width must satisfy $0.85 \epsilon \leq e \leq 1.25 \epsilon$, where $\epsilon$ is the width of the space character of the used font.
In this case, the cost of an arc is the width of a word in its font. Because we want three lines, we have to check three different sums (one for each line). To do so, each time the sum associated with a line is valid, the sum is reset to 0.

Once computed, MDDMNREAD contains sentences that satisfy all rules  $(\Rule{i})_{i=0..4}$. 

%\begin{figure}
%    \centering
%   \includegraphics[width=0.3\textwidth]{ima/firstline.png}
%    \caption{A figure that shows that, The white cat loves is a valid assignement to fullfill first line relative to display constraint }
%    \label{fig:firstline}
%\end{figure}

\subsection{Sentences Selection by LM}
Unfortunately, \ngrams only partially handle the meaning and grammatical concerns of the sentences. Indeed they have limitations, especially in the case of long-range dependencies (i.e., where the dependency size between two words exceed the \ngrams size).
Therefore, we propose to use an LM to select candidate sentences (correct sentences with meaning) inside the solutions space. 
\subsubsection{Model Choice: a Generative Model}
There are a lot of transformers-based LM due mainly to their effectiveness. 
These large LMs are trained on several tens of gigabytes of text. These texts are cut into chunks of thousands tokens, and LMs process these sequences of tokens to learn a text distribution. Once trained, generative models (like GPT-2) can generate sequences using a part of existing sequences or a token-of-start to compute a distribution of potential successors.
Thus, one of the main goals of an LM is the capability to compute the probability of a sequence based on training data.
We choose to use GPT-2 because it is recognized as one of the best in language modeling tasks on several benchmarks\footnote{https://paperswithcode.com/paper/language-models-are-unsupervised-multitask\#results}.
\subsubsection{Sentences Scoring: Perplexity}
As a matter of fact, from GPT-2, which computes the probability of a sequence, we can compute a perplexity score of that sequence.
Perplexity is an entropy metric derived from Shannon's information theory \cite{brown-etal:1992}. 
It can be expressed as the geometric mean of the inverse conditional likelihood of the sequence \cite{jurafsky:2009}. Given $S_n$ the sequence of a succession of words of size $n$, so $S_n = w_1w_2..w_n$. The perplexity (PPL) of $S_n$ is computed as follows:
\[ PPL(S_n) = \sqrt[n]{\frac{1}{P(w_1w_2w_3...w_n)}},\]
where probability $P(\cdot)$ is given by the LM. 
PPL can be interpreted as the "uncertainty" of the model with respect to a sample.
Usually, it is used to evaluate the LM itself by checking that good samples are recognized as such (i.e., low PPL values). 
\subsubsection{GPT-2 as an Oracle}
%We propose to use the perplexity to differentiate between good and bad sentences.
%Ideally we would like GPT-2 to be able to automatically differentiate between good and bad sentences. 
%One way to evaluate a sample with a LM is to compute the perplexity of the sequences with respect with the LM.
%Here we do a kind of hypothesis reversal. 
We suppose that GPT-2 has grasped a part of the general language  and that its score is trustworthy. Thus, we compute the perplexity of the sentences as a score. This score will assess the correctness and fluency of the sentences. 
So, if GPT-2 assigns a low PPL to a sentence, we conclude that it is very likely to be a "good" sentence.
By "good" sentence, we mean a grammatically correct sentence that makes sense.
Once the sentences are generated, we import a pre-trained model of the language of the sentences generated and we sort them according to the PPL score.

%\paragraph{How do we score sentences without knowing that we are in constrained spaces ?}
%Using an independent language model; Even if part of theses models are used to generate text, we can ask them to evaluate a sentence. More precisely, this is a way of asking the model of transformers how it evaluates that it could have generated this sentence. Let's use PPL.
%But we can reverse these assumptions and claim that the model has captured part of the general distribution of language. 
%Thus we can score sentences based on the distribution learn by GPT-2. 

%\include{inc/algo}

%\subsection{Operations between MDDs: calculation of solutions} 

%To obtain solutions that satisfy all the constraints, the final MDD is calculated as the on-the-fly intersection of the MDDs defined previously. This on-the-fly intersection operation and all the other operations on the MDDs are based on an algorithm associated with the generic operator APPLY $\bigoplus$ defined in~\cite{perez-regin:15,jung-regin:22}. \color{blue} The whole point of this on-the-fly intersection is to make it possible to carry out this constraint conjunction operation efficiently on the MDDs without building the righmost MDD. \color{black}

\section{Results}
\label{sec:resultats}

\subsection{Experimental Conditions}
The model described in Sec. \ref{sec:approach} was implemented in Java 17.
The code is available upon request. 
The experiments were performed on a machine using a Intel(R) Xeon(R) W-2175 CPU @ 2.50GHz with 256 GB of RAM and running under Ubuntu 18.04. 

The sentence selection task was performed with models without any fine-tuning by using either OpenAI GPT-2 \cite{radford:2019} for English sentences or a French trained GPT-2 \cite{simoulin:2021}. These models are available from the huggingface library \cite{hugginface:2020}.

\subsubsection{Target Language}
Our primary concern is the French language. It is a hard Latin language. We also generate sentences in English as a proof-of-concept of the multilingual aspect of our approach.

\subsubsection{Corpus of N-grams}
The implications of corpus construction are developed in Sec. \ref{corpus:babling}.

For French, to build our corpus of \ngrams, we started with 443 books belonging to the youth category. This choice was motivated by generating sentences having a simple structure and using a simple lexicon (see $\regleLexique$)

For English, to constitute our corpus, we build a set of 75 books fiction category. Nevertheless, we did not “fine-tune” our corpus as much as for French since our goal was obviously to show the multilingual potential of our approach than to produce the best sentences in English.

\subsection{Performance Analysis}
Table \ref{table:perf} summarizes MDDs computational data.
Our model is fast. Some effort was made in this direction for responsiveness.
MDDtrie requires a non-negligible amount of memory to be computed (30GB). However, it is deeply related to the fact that strings are generally memory intensive and especially in Java.
Nevertheless, since MDDtrie is almost a form of preprocessing of the data input, we do not think that is a critical part for the moment.

\begin{table}[htbp]  
\resizebox{\textwidth}{!}{
    \begin{tabular}{|l|r|r|r|r|r|}
\hline
MDD & arcs	& nodes	& solutions	& GB & s  \\
    \hline
MDDT &	4,972,698 &	1,171,904 &	3,981,618 & 30 & 66 \\
MDDM &	23,943 &	18,983 &	7,028 &	3 & 72\\ 
\hline
MDDT &	981,372 &	268,291 & 735,928 &	6 & 8 \\
MDDM &	 1390 &	1201 &	204 &	$\ll 1$ & 3 \\
\hline
\end{tabular}}

\caption{\label{table:perf}Number of arcs, nodes, solutions, gigabytes (GB), and seconds (s) for computing MDDTrie (MMDT) and MDDMNREAD (MDDM) from 5-grams. Lines 2 and 3 are for French and 4 and 5 for English.}
\end{table}

\begin{table}[htbp]
\centering
\small
\resizebox{\textwidth}{!}{
\begin{tabular}{|l|l|}
\hline
English Generated Sentences & PPL \\
\hline
And he had no idea what to do with the fact that she was in  &17.9 \\
It made me wonder if it was going to be able to do the work  &20.1 \\
You should be able to get out of bed to get out of the room  &20.2 \\
You need to get out of bed to get out of the room right now  &21.2\\
No one will be able to get in and out of the room right now  &21.4 \\
.... & ... \\
The family in front of the double doors and into the branch  &109 \\
She paused outside the door to the back of my left shoulder     &123 \\
The difference here was now she had to deal with in my life    &160 \\
She hesitated at the door to the back of her nose and mouth    &189 \\
So strange to be on the edge of my chin between his fingers    & 224 \\

\hline
\end{tabular}}
\caption{\label{tab:sentences_scoring_en} Examples of generated English sentences sorted by PPL score using gpt-2 \tablefootnote{https://huggingface.co/gpt2}
}
%\end{table}
%
\ \\ 
%\begin{table}[htbp]
\centering
\small
\resizebox{\textwidth}{!}{
\begin{tabular}{|l|l|}
\hline
French Generated Sentences & PPL \\
\hline
Assise à la table de la salle à manger il y a un bon moment & 6.23 \\
Elle se pencha vers elle et lui donna un grand coup de pied & 6.62 \\
Et il y a des choses dont on ne peut pas dire la même chose & 6.82 \\
Nous avons à peine le temps de faire un tour dans la maison & 6.83 \\
Ses parents ne sont pas au courant de la maladie de sa mère & 6.94 \\
.... & ... \\
Il frappa à nouveau dans ses yeux et de voir la grue bouger & 230 \\
La poupée finit par tomber sur le sol et préparai mon bâton & 241 \\
Une hache reposait près de lui et de ne pas jouer aux héros & 241 \\
Quand le bout de la salle et frappa dans ses mains en coupe & 242 \\
Une hache reposait près de lui et de ne pas sortir ensemble & 291 \\
\hline
\end{tabular}}
\caption{\label{tab:sentences_scoring_fr} Examples of French generated sentences sorted by PPL score using asi/gpt-fr-cased-base\tablefootnote{https://huggingface.co/asi/gpt-fr-cased-base}
}
\ \\ 
 \centering
 \resizebox{\textwidth}{!}{
     \begin{tabular}{|l|l|}
     \hline
Sentences & PPL \\
\hline
Le policier passa la main dans sa poche et se mit à marcher &23 \\
Le policier passa la main dans sa poche et se mit à compter &26 \\
Le policier passa la main dans la poche arrière de mon jean &38 \\
Le policier passa la main dans la poche de son jean déchiré &42 \\
Le policier passa la main dans la poche de sa nouvelle robe &76\\
        \hline
        
     \end{tabular}}
    \caption{\label{tab:exemple-de-phrases-avec-meme-amorce} Examples of sentences generated in 5-grams with the same sentence start.}
\end{table}

\subsection{Sentence Analysis}
Tables \ref{tab:sentences_scoring_en} and \ref{tab:sentences_scoring_fr} show a sample of some sorted sentences computed from the solution set generated by MDDMNREAD. This sample includes the lowest and largest PPL values. Interestingly, PPL scoring succeeds in sorting sentences.
Excellent sentences can be found in the first positions, and very bad sentences can be found in the last positions.

Identifying multiple threshold values, each of which delineates a range within the set of solutions is not straightforward.
We observe that we have perfect sentences between the lowest PPL values and PPL less than 15. Between 15 and 30, it gets complicated with good sentences and sometimes few strange sentences. For PPL values greater than 30, we still find some excellent sentences, but we need to become cautious, so it is as if we were selecting sentences without help. 

Since the PPL range is $ [1 ; +\infty ]$  it is harder to give a confidence interval for the last position because the span between the two last values can be greater than the span between the first sentence and the before-last sentence valuation.

Also, PPL score is strongly dependent on the vocabulary and \emph{a fortiori} the LM (i.e., training data). Therefore with another LM, we have no guarantee that the PPL min/max and, more generally, the valuation would be similar.

Finally, PPL valuation through GPT-2 is slow. French sentences scoring takes roughly 1 hour.

%\subsection{Sentence analysis}
%Dit autre part 
%Among sentences we generate, there are candidate sentences, i.e., whose syntax and meaning are correct, but also sentences whose meaning is curious or sentences whose syntax is not correct %(see Tab.~\ref{tab:types-de-phrases}).

%the best ranked sentences are good. All sentences are good up to this value.
 %Between v1 and v2, many sentences are good but there are also some doubtful sentence
 %The lowest ranked sentences are bad. All the sentences are bad from such value.
 
The system tends to generate all the variants with the same sentence start (see Tab.~\ref{tab:exemple-de-phrases-avec-meme-amorce}). These variants can all be of good quality, as seen with the beginning: 
\emph{``Le policier passa la main dans sa poche"}. However, since we consider the PPL score, only two first sentences would be considered for MNREAD use. PPL helps to choose the best sentences between variants of the sentences.
%However, to form a test set of sentences, note that we must ensure variability to avoid having similar bits of sentences in the same set.

%\begin{table}[hbtp]
   
%\end{table}

%Another well-known observation that can be made is that the smaller the size of \ngram used to build the sentences, the lower the grammatical and semantic quality of the sentences. That is why we choose to show 5-grams benchmarks.

%With the sorting using the PPL, one could decrease the size of the n-grams and thus obtain more sentences.
%Decreasing the size to 4-grams is still possible, although the combinatorics increases strongly. The sorting capacity of GPT-2 would also be affected by some of these errors (i.e., more bad sentences with a good score).
%To conclude, keep high n-grams sizes (4 or 5) and, according to the improvement of the future LM, decrease towards smaller n-grams according to the needs, but be careful with the skyrocketing combinatorics of smaller \ngrams.

An established observation in NLP is that lower-size n-grams used to construct text tend to result in a lower grammatical and semantic quality of the output. In this paper, we choose to present benchmarks based on 5-grams. It is possible to decrease the n-gram size further to 4-grams, although this would substantially increase combinatorics. Furthermore, 
we observe that sometimes good sentences receive poor scores and vice versa. Reducing the n-gram size would result in a way higher occurrence of this phenomenon.

In conclusion, using higher n-gram sizes (4 or 5) to construct sentences is advisable. However, as LMs performances continuously improve, it may become possible to decrease the n-gram size according to the requirements of specific tasks. Nevertheless, be careful when considering lower-size n-grams due to their skyrocketing combinatorics.

\subsection{Sentence Evaluation}

The purpose of MNREAD sentences is to be read at the same speed. We want to make sure that the generated sentences are equivalent to the original ones. This sentence evaluation allows us to validate our approach, particularly the use of the PPL.

\subsubsection{Experimental Design} We asked 14 normally sighted participants (age 14 to 56) to read 3 sets of 30 French sentences: MNREAD sentences, generated sentences ranked as ‘good’ (denoted by low\_PPL) and generated sentences ranked as ‘bad’ (denoted by high\_PPL). All sentences were displayed at 40cm, in regular polarity, with a fixed print size. %using a virtual reality version of MNREAD. 
Corrected reading speed was measured in words/min (wpm) and analyzed using a mixed-effects model.%(see Fig. \ref{fig:sujet11}).

\subsubsection{Statistical Analysis} 
\begin{figure}
    \centering
    \includegraphics[width=0.5\textwidth]{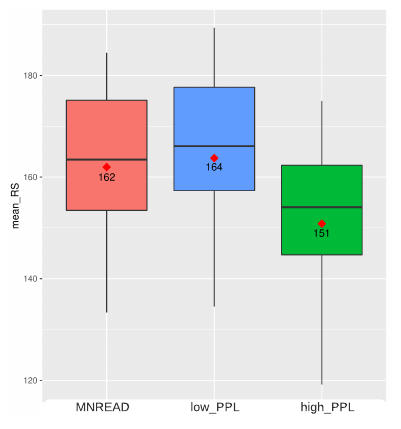}
    \caption{Box plot that shows Mean Reading Speed (MRS) measured for three sentences set “MNREAD”,“low\_PPL”, “high\_PPL”.}
    \label{fig:boxplot}
\end{figure}

The results are shown in the  Fig. \ref{fig:boxplot}.
On average, ‘low\_PPL’ generated sentences were read at 164 wpm. This value was not significantly different from the reading speed of 162 wpm yield by MNREAD sentences ($p=0.5$). On the other hand, reading speed was significantly slower for ‘high\_PPL’ generated sentences, with an average value of 151 wpm ($p<0.001$). Therefore, this preliminary study suggests that our method provides valid standardized sentences that follow the MNREAD rules and yield similar performance, at least in French. %Further investigations are needed to validate the generation of sentences in English, Spanish, Italian, etc. 

PPL helps in two ways:
(i) First, we were able to select good sentences for the MNREAD application.
(ii) As a side effect, PPL may be a good predictor of reading speed. Therefore, high PPL will tend us to think that the sentences are not suited for the test because their reading speed may be lower than canonical ones.

%\begin{figure}[htbp]
%    \centering
%    \includegraphics[width=0.5\textwidth]{ima/validation_phrases_sujet11.pdf}
%    \caption{Caption}
%    \label{fig:sujet11}
%\end{figure}
\section{Discussion}
\label{sec:discussion}
\subsection{Post-generation Selection}

%In the results section, we mention that we have sentences that do not make sense in the set of feasible solutions. The gold standard evaluation for sentence (text) generation is human evaluation. But even if we limit our solution set to a few thousand sentences ;
%Reading thousands of sentences remains a tedious task. At this stage, we want an automatic sentence evaluation that, in the best case, avoids human verification or at least gives a confidence interval on the excellent candidate solutions.
Since we use \ngrams in our approach, our first idea was to build the distribution of the n-gram language model \cite{jurafsky:2009} from the empirical occurrences of words in the input corpus (i.e., perform Markov sampling on the solution set). Many works have been done in this spirit, especially in music generation \cite{pachet-roy:2011}.
It gave better results than a random selection but lacked robustness. The drawback of Markov sampling is that it evaluates the meaning of a sentence only from the constrained solution set, whereas it should consider the whole language instead. 

%The following question then arises: How do we score sentences without knowing that we are in constrained spaces? Using an independent language model;
%Although the Transformers model cannot be used directly to solve this task;
%These models, in particular GPT-2, have shown impressive results in text generation.
%Even if these models are used to generate text, we can ask them to evaluate a sentence. More precisely, this is a way of asking the model of transformers how it evaluates that it could have generated this sentence. To do this, we use the well-known perplexity metric.

%Usually, it is used to evaluate the language of the model itself by choosing a good sample and validating that the good sample is recognized as such. But we can reverse these assumptions and claim that the model has captured part of the general distribution of language. 
%Thus it can score sentences based on the distribution learn by GPT-2. 

We choose to evaluate solutions after the generation stage for two reasons: 
(i) The transformer's valuation is time-consuming due to the quadratic complexity architecture. Even though some work is currently done to build a new architecture to reduce its algorithmic complexity (e.g., forecasting time series \cite{informeraaai:2021}). 
(ii) The perplexity is the geometric mean of the inverse conditional probability of the sequences.
It means that we need the entire sentence to obtain the exact valuation. This also means that if we want to score all subsequences (subsentences), it will produce a tremendous amount of requests to the LM at the generation stage because millions of them are produced.
Nevertheless, we used it as a powerful oracle for scoring generated sentences.

\subsection{CP-ML Hybridization: Perplexity as a Constraint}
Currently, the perplexity computed by an LM is not integrated into the model as an additional constraint.
This is not an issue because the number of generated sentences is in the order of a few hundreds or thousands and, therefore, not too large. 
However, millions of sentences could be generated if the constraints were relaxed or if the input corpus size were significantly increased.
Evaluating millions of sentences, even on a powerful machine with GPT-2, can take several weeks.

To define a constraint of perplexity $PPL( X < K)$, it is necessary to be able to decide that for a partial assignment of variables (e.g., the variables from 1 to $k$), one can ensure that the total perplexity (for the variables from 1 to $n$) will necessarily exceed $K$. Thus, in this case, we will reject the partial assignment by not continuing it. The major problem is that the perplexity involves all the variables, and the relation between the perplexity of a subset of variables and the full set of variables is not clear. It is not monotonic and can vary greatly and, in particular, improve significantly. Therefore, designing a perplexity constraint is not straightforward.

\color{blue}

\color{black}
\subsection{Genericity of the Method}

A more ad-hoc approach could have been used. However, thanks to MDD, particularly the second one, our method can be easily generalized to integrate other constraints or deal with similar problems. (i) Additional constraints can be integrated similarly to those we have considered (e.g., all-diff, gcc),
i.e., during the construction of the final MDD from MDDtrie. (ii) We can be even more generic by proceeding by intersections of MDD. Each constraint is associated with an MDD. Then they are integrated with the unfolded MDDtrie by intersecting the MDDs. Our first model %\cite{bonlarron2022generation} 
was originally implemented in this way. This alternative is highly modular and generic because we can obtain, store and reuse any intermediate results. However, it is less efficient in time and memory (counterintuitively, the intersection of two MDDs can take more memory than the two original MDDs because of the local decompressions that can arise). For instance, we needed 64 GB and 6 hours to run the experiments instead of 32 GB and 2 minutes with the current approach.

\subsection{The Choice of the Corpus }
\label{corpus:babling}

The choice of the corpus is not critical but deserves particular attention because it strongly impacts the kind of generated sentences.

For example, we tested Wikipedia as an input corpus. The sentences' tone appears to be "cold", very impersonal and descriptive (like \emph{"This was one of the most famous in the history of the sport"} or \emph{"It may also be used in the same way as the rest of the army"} or \emph{"Finalement le cœur de la ville est située au nord et au sud"}). 
This comes from the fact that our method is based on \ngrams and \ngrams tend to produce text in the style of an author \cite{papadopoulos-roy-pachet:2014} and because Wikipedia is an encyclopedia. These sentences have not been considered well suited for a vision test.

%Les google ngram ont une vocation à représenter une grande partie du texte ecrit dans une langue avec tout type de support. Cela crée des problems car certaines phrases vont engendrer de nombreux ngram peut acceptable pour notre utilisation. Par exemple

%Des filtres ad-hoc et non évidents doivent etre definis et appliquée sur une quantité gigantesque de données (terabytes). La tache s'avere beaucoup plus complexe qu'il n'y parait.
Another issue was found when we tested a subpart of Google \ngrams corpus \cite{michel-etal:11}.
The Google \ngrams represent a large part of the text written in several languages with a wide variety of sources (Google books database). This creates problems because 
some sentences will lead to the production of \ngrams which are not acceptable for our use (e.g., bad \ngrams of end of sentences). For instance : \emph{"- You... you're not allowed... Dad told you that you shouldn't do ma... magic..."}.
%"Papa t’a dit que tu ne devais pas faire de ma… de magie." va donner un ngram qui termine sur ou qui autorise "ma de"
Ad-hoc and non-obvious filters must be defined and applied to a considerable amount of data (terabytes). The task is much more complex than it seems.

For all of these reasons, we chose to remain on books for the moment.
However, books must also be chosen carefully, because word occurrences are essential. 
For instance, consider the word "\emph{tomato}". Suppose there are not enough books that tell stories about "\emph{tomato}" in the input corpus. Then the model cannot produce sentences containing the word "\emph{tomato}". 
This observation was made when we put some books about french history. As soon as we did it, we generated sentences with the words "\emph{king}", "\emph{queen}", "\emph{castle}"... 
When creating the corpus, we should therefore consider all the books and the possible relations between them.
To go even further in the reasoning. Is it reasonable to mix n-grams from two very far topics like science fiction and heroic fantasy books, allowing the emergence of sentences mentioning "\emph{dragon knight flying in the depths of the universe}"?

To conclude this section, the choice of the corpus seems to be more an artistic choice than an methodological one.

\section{Conclusion}
\label{sec:conclusion}

A novel and creative approach for constrained text generation has been presented. It formalizes the problem as a discrete combinatorial optimization problem and solves it using a model based on MDDs.
We introduce MNREAD sentence problem generation as a constrained text generation task and we successfully applied our method, allowing the creation of new sets of sentences for MNREAD test. 
%Hence, CP approach remains competitive for constrained text generation task by producing thousands of MNREAD sentences. 
We have generated thousands of sentences in French and hundreds of sentences in English in roughly two minutes. These results will lead to broader use of the MNREAD test. 
In addition, we show that GPT-2 can be used to cleverly sort the generated sentences to simplify human verification of generated sentences. Our general method can generate any standardized text content, especially with hard constraints.
It also brings an outside-the-box point of view on the constrained text generation task by solving the constraints first.

\section*{Acknowledgments}
This work has been supported by the French government, through the 3IA C\^ote d'Azur Investments in the
Future project managed by the National Research Agency (ANR) with the reference number ANR-19-P3IA-0002.

\bibliographystyle{unsrt}
\bibliography{main}

\begin{thebibliography}{10}

\bibitem{glue:2018}
Alex Wang, Amanpreet Singh, Julian Michael, Felix Hill, Omer Levy, and
  Samuel~R. Bowman.
\newblock {GLUE:} {A} multi-task benchmark and analysis platform for natural
  language understanding.
\newblock {\em CoRR}, abs/1804.07461, 2018.

\bibitem{roush-etal-2022-language}
Allen Roush, Sanjay Basu, Akshay Moorthy, and Dmitry Dubovoy.
\newblock Most language models can be poets too: An {AI} writing assistant and
  constrained text generation studio.
\newblock In {\em Proceedings of the Second Workshop on When Creative AI Meets
  Conversational AI}, pages 9--15, Gyeongju, Republic of Korea, October 2022.
  Association for Computational Linguistics.

\bibitem{beamsearch1:2021}
Yixian Liu, Liwen Zhang, Wenjuan Han, Yue Zhang, and Kewei Tu.
\newblock Constrained text generation with global guidance - case study on
  commongen.
\newblock {\em CoRR}, abs/2103.07170, 2021.

\bibitem{beamsearch2:2018}
Matt Post and David Vilar.
\newblock Fast lexically constrained decoding with dynamic beam allocation for
  neural machine translation.
\newblock In {\em Proceedings of the 2018 Conference of the North {A}merican
  Chapter of the Association for Computational Linguistics: Human Language
  Technologies, Volume 1 (Long Papers)}, pages 1314--1324, New Orleans,
  Louisiana, June 2018. Association for Computational Linguistics.

\bibitem{hokamp-liu:2017}
Chris Hokamp and Qun Liu.
\newblock Lexically constrained decoding for sequence generation using grid
  beam search.
\newblock In {\em Proceedings of the 55th Annual Meeting of the Association for
  Computational Linguistics (Volume 1: Long Papers)}, pages 1535--1546,
  Vancouver, Canada, July 2017. Association for Computational Linguistics.

\bibitem{lu-etal-2022-neurologic}
Ximing Lu, Sean Welleck, Peter West, Liwei Jiang, Jungo Kasai, Daniel Khashabi,
  Ronan Le~Bras, Lianhui Qin, Youngjae Yu, Rowan Zellers, Noah~A. Smith, and
  Yejin Choi.
\newblock {N}euro{L}ogic {A}*esque decoding: Constrained text generation with
  lookahead heuristics.
\newblock In {\em Proceedings of the 2022 Conference of the North American
  Chapter of the Association for Computational Linguistics: Human Language
  Technologies}, pages 780--799, Seattle, United States, July 2022. Association
  for Computational Linguistics.

\bibitem{papadopoulos:2015}
Alexandre Papadopoulos, Pierre Roy, Jean-Charles R{\'e}gin, and Fran{\c c}ois
  Pachet.
\newblock {Generating all Possible Palindromes from Ngram Corpora}.
\newblock In {\em {IJCAI 2015}}, Buenos Aires, Argentina, July 2015.

\bibitem{mansfield-ahn-etal:93}
J.~Stephen Mansfield, Sonia~J. Ahn, Gordon~E. Legge, and Andrew Luebker.
\newblock A new reading-acuity chart for normal and low vision.
\newblock {\em Ophthalmic and Visual Optics/Noninvasive Assessment of the
  Visual System Technical Digest}, 3:232–235, 1993.

\bibitem{vasmani-et-al:2017}
Ashish Vaswani, Noam Shazeer, Niki Parmar, Jakob Uszkoreit, Llion Jones,
  Aidan~N Gomez, {\L}ukasz Kaiser, and Illia Polosukhin.
\newblock Attention is all you need.
\newblock In {\em Advances in Neural Information Processing Systems}, pages
  5998--6008, 2017.

\bibitem{radford:2019}
Tom Brown, Benjamin Mann, Nick Ryder, Melanie Subbiah, Jared~D Kaplan, Prafulla
  Dhariwal, Arvind Neelakantan, Pranav Shyam, Girish Sastry, Amanda Askell,
  et~al.
\newblock Language models are few-shot learners.
\newblock {\em Advances in neural information processing systems},
  33:1877--1901, 2020.

\bibitem{calabrese-owsley-etal:16}
Aurélie Calabrese, Cynthia Owsley, Gerald McGwin, and Gordon~E. Legge.
\newblock Development of a reading accessibility index using the {MNREAD}
  acuity chart.
\newblock {\em JAMA Ophthalmol.}, 134(4):398–405, April 2016.

\bibitem{crossland-legge-etal:07}
Michael~D. Crossland, Gordon~E. Legge, and Steven~C. Dakin.
\newblock The development of an automated sentence generator for the assessment
  of reading speed.
\newblock {\em Behavioral and Brain Functions}, 4(1):14, 2008.

\bibitem{perrin-paille-etal:15}
Jean-Luc Perrin, Damien Paill{\'e}, and Thierry Baccino.
\newblock A new sentence generator providing material for maximum reading speed
  measurement.
\newblock {\em Behav Res}, 47:055–1064, 2015.

\bibitem{mansfield-atilgan-etal:19}
J.~Stephen Mansfield, Nilsu Atilgan, Angela Lewis, and Gordon Legge.
\newblock Extending the {MNREAD} sentence corpus: Computer-generated sentences
  for measuring visual performance in reading.
\newblock {\em Vision research}, 158:11--18, 2019.

\bibitem{lete-sprenger-charolles-etal:04}
Bernard Lété, Liliane Sprenger-Charolles, and Pascale Colé.
\newblock Manulex : A grade-level lexical database from french
  elementary-school readers.
\newblock {\em Behavior Research Methods, Instruments \& Computers},
  36:166–176, 2004.

\bibitem{akers:78}
Sheldon~B. Akers.
\newblock Binary decision diagrams.
\newblock {\em IEEE Transactions on Computers}, C(27):509–516, June 1978.

\bibitem{roy-perez-etal:16}
Pierre Roy, Guillaume Perez, Jean-Charles Régin, Alexandre Papadopoulos,
  F.~Pachet, and M.~Marchini.
\newblock Enforcing structure on temporal sequences: the {A}llen constraint.
\newblock In {\em International conference on principles and practice of
  constraint programming}, page 786–801. Springer, 2016.

\bibitem{perez-regin:17b}
Guillaume Perez and Jean-Charles R{\'e}gin.
\newblock {MDDs}: Sampling and probability constraints.
\newblock In {\em Proceedings of the International Conference on Principles and
  Practice of Constraint Programming}, page 226–242, 2017.

\bibitem{cheng-yap:10}
Kenil~C.K. Cheng and Roland~H.C. Yap.
\newblock An {MDD}-based generalized arc consistency algorithm for positive and
  negative table constraints and some global constraints.
\newblock {\em Constraints}, 15:265–304, 2010.

\bibitem{lecoutre:2011}
Christophe Lecoutre.
\newblock {STR2}: Optimized simple tabular reduction for table constraints.
\newblock {\em Constraints}, 16(4):341–371, oct 2011.

\bibitem{Wang-Yap:2022}
Ruiwei Wang and Roland~H.C. Yap.
\newblock Encoding multi-valued decision diagram constraints as binary
  constraint trees.
\newblock {\em Proceedings of the AAAI Conference on Artificial Intelligence},
  36(4):3850--3858, Jun. 2022.

\bibitem{gentzel-etal:2022}
Rebecca Gentzel, Laurent Michel, and Willem-Jan van Hoeve.
\newblock {Heuristics for MDD Propagation in HADDOCK}.
\newblock In Christine Solnon, editor, {\em 28th International Conference on
  Principles and Practice of Constraint Programming (CP 2022)}, volume 235 of
  {\em Leibniz International Proceedings in Informatics (LIPIcs)}, pages
  24:1--24:17, Dagstuhl, Germany, 2022. Schloss Dagstuhl -- Leibniz-Zentrum
  f{\"u}r Informatik.

\bibitem{verhaeghe2019extending}
H{\'e}l{\`e}ne Verhaeghe, Christophe Lecoutre, and Pierre Schaus.
\newblock Extending compact-diagram to basic smart multi-valued variable
  diagrams.
\newblock In {\em Integration of Constraint Programming, Artificial
  Intelligence, and Operations Research: 16th International Conference, CPAIOR
  2019, Thessaloniki, Greece, June 4--7, 2019, Proceedings}, pages 581--598.
  Springer, 2019.

\bibitem{verhaeghe2018compact}
H{\'e}l{\`e}ne Verhaeghe, Christophe Lecoutre, and Pierre Schaus.
\newblock Compact-mdd: Efficiently filtering (s) {MDD} constraints with
  reversible sparse bit-sets.
\newblock In {\em Twenty-Seventh International Joint Conference on Artificial
  Intelligence {IJCAI-18}}, 2018.

\bibitem{bergman-cire-etal:16}
David Bergman, Andre~A. Cire, Willem-Jan Hoeve, and John Hooker.
\newblock {\em Decision Diagrams for Optimization}.
\newblock Springer Publishing Company, Incorporated, 1st edition, 2016.

\bibitem{perez:17}
Guillaume Perez.
\newblock {\em Diagrammes de décision : contraintes et algorithmes}.
\newblock PhD thesis, Université Côte d'Azur, 2017.

\bibitem{gillard2022large}
Xavier Gillard and Pierre Schaus.
\newblock Large neighborhood search with decision diagrams.
\newblock In {\em International Joint Conference on Artificial Intelligence},
  2022.

\bibitem{rudich:2022peelandbound}
Isaac Rudich, Quentin Cappart, and Louis{-}Martin Rousseau.
\newblock Peel-and-bound: Generating stronger relaxed bounds with multivalued
  decision diagrams.
\newblock In Christine Solnon, editor, {\em 28th International Conference on
  Principles and Practice of Constraint Programming, {CP} 2022, July 31 to
  August 8, 2022, Haifa, Israel}, volume 235 of {\em LIPIcs}, pages
  35:1--35:20. Schloss Dagstuhl - Leibniz-Zentrum f{\"{u}}r Informatik, 2022.

\bibitem{Hoeve:2022}
Willem-Jan van Hoeve.
\newblock Graph coloring with decision diagrams.
\newblock {\em Mathematical Programming}, 192(1):631--674, 2022.

\bibitem{Amilhastre:2014}
J\'{e}r\^{o}me Amilhastre, H\'{e}l\`{e}ne Fargier, Alexandre Niveau, and
  C\'{e}dric Pralet.
\newblock Compiling csps: A complexity map of (non-deterministic) multivalued
  decision diagrams.
\newblock {\em International Journal on Artificial Intelligence Tools},
  23(04):1460015, 2014.

\bibitem{schmid:13}
Helmut Schmid.
\newblock Probabilistic part-of-speech tagging using decision trees.
\newblock In {\em New methods in language processing}, page 154, 2013.

\bibitem{perez-regin:15}
Guillaume Perez and Jean-Charles R{\'e}gin.
\newblock Efficient operations on mdds for building constraint programming
  models.
\newblock In {\em Proceedings of the 24th International Conference on
  Artificial Intelligence}, pages 374--380, 2015.

\bibitem{jung-regin:22}
Victor Jung and Jean-Charles R{\'e}gin.
\newblock Efficient operations between mdds and constraints.
\newblock In Pierre Schaus, editor, {\em Integration of Constraint Programming,
  Artificial Intelligence, and Operations Research}, pages 173--189, Cham,
  2022. Springer International Publishing.

\bibitem{trick:03}
Michael Trick.
\newblock A dynamic programming approach for consistency and propagation for
  knapsack constraints.
\newblock {\em Annals of Operations Research}, 118:73–84, 2003.

\bibitem{brown-etal:1992}
Peter~F. Brown, Stephen~A. Della~Pietra, Vincent~J. Della~Pietra, Jennifer~C.
  Lai, and Robert~L. Mercer.
\newblock An estimate of an upper bound for the entropy of {E}nglish.
\newblock {\em Computational Linguistics}, 18(1):31--40, 1992.

\bibitem{jurafsky:2009}
Dan Jurafsky and James~H. Martin.
\newblock {\em Speech and language processing : an introduction to natural
  language processing, computational linguistics, and speech recognition}.
\newblock Pearson Prentice Hall, Upper Saddle River, N.J., 2009.

\bibitem{simoulin:2021}
Antoine Simoulin and Benoit Crabb{\'e}.
\newblock {Un mod{\`e}le Transformer G{\'e}n{\'e}ratif Pr{\'e}-entrain{\'e}
  pour le fran{\c c}ais}.
\newblock In Pascal Denis, Natalia Grabar, Amel Fraisse, R{\'e}mi Cardon,
  Bernard Jacquemin, Eric Kergosien, and Antonio Balvet, editors, {\em
  {Traitement Automatique des Langues Naturelles}}, pages 246--255, Lille,
  France, 2021. {ATALA}.

\bibitem{hugginface:2020}
Thomas Wolf, Lysandre Debut, Victor Sanh, Julien Chaumond, Clement Delangue,
  Anthony Moi, Pierric Cistac, Tim Rault, Rémi Louf, Morgan Funtowicz, Joe
  Davison, Sam Shleifer, Patrick von Platen, Clara Ma, Yacine Jernite, Julien
  Plu, Canwen Xu, Teven~Le Scao, Sylvain Gugger, Mariama Drame, Quentin Lhoest,
  and Alexander~M. Rush.
\newblock Transformers: State-of-the-art natural language processing.
\newblock In {\em Proceedings of the 2020 Conference on Empirical Methods in
  Natural Language Processing: System Demonstrations}, Online, October 2020.
  Association for Computational Linguistics.

\bibitem{pachet-roy:2011}
Fran\c{c}ois Pachet and Pierre Roy.
\newblock Markov constraints: Steerable generation of markov sequences.
\newblock {\em Constraints}, 16(2):148–172, apr 2011.

\bibitem{informeraaai:2021}
Haoyi Zhou, Shanghang Zhang, Jieqi Peng, Shuai Zhang, Jianxin Li, Hui Xiong,
  and Wancai Zhang.
\newblock Informer: Beyond efficient transformer for long sequence time-series
  forecasting.
\newblock {\em Proceedings of the AAAI Conference on Artificial Intelligence},
  35(12):11106--11115, May 2021.

\bibitem{papadopoulos-roy-pachet:2014}
Alexandre Papadopoulos, Pierre Roy, and Fran\c{c}ois Pachet.
\newblock Avoiding plagiarism in markov sequence generation.
\newblock In {\em Proceedings of the Twenty-Eighth AAAI Conference on
  Artificial Intelligence}, AAAI'14, page 2731–2737. AAAI Press, 2014.

\bibitem{michel-etal:11}
Jean-Baptiste Michel, Yuan~Kui Shen, Aviva~Presser Aiden, Adrian Veres,
  Matthew~K. Gray, The Google~Books Team, Joseph~P. Pickett, Dale Hoiberg, Dan
  Clancy, Peter Norvig, Jon Orwant, Steven Pinker, Martin~A. Nowak, and
  Erez~Lieberman Aiden.
\newblock Quantitative analysis of culture using millions of digitized books.
\newblock {\em Science}, 331(6014):176--182, 2011.

\end{thebibliography}

\end{document}